# Directed Graph Attention Neural Network Utilizing 3D Coordinates for Molecular Property Prediction

*Chen Qian,*[*a] *Yunhai Xiong, Xiang Chen*[*b]

a Department of Mechanical Engineering, Zhejiang University, 866 Yuhangtang Rd, Hangzhou 310058, PR China; Email: 11525069@zju.edu.cn

b MIIT Key Laboratory of Advanced Display Materials and Devices, Institute of Optoelectronics & Nanomaterials, School of Materials Science and Engineering, Nanjing University of Science and Technology, Nanjing 210094, China; Email: xiangchen@njust.edu.cn

ABSTRACT: The prosperity of computer vision (CV) and natural language procession (NLP) in recent years has spurred the development of deep learning in many other domains. The advancement in machine learning provides us with an alternative option besides the computationally expensive density functional theories (DFT). Kernel method and graph neural networks have been widely studied as two mainstream methods for property prediction. The promising graph neural networks have achieved comparable accuracy to the DFT method for specific objects in the recent study. However, most of the graph neural networks with high precision so far require fully connected graphs with pairwise distance distribution as edge information. In this work, we shed light on the Directed Graph Attention Neural Network (DGANN), which only takes chemical bonds as edges and operates on bonds and atoms of molecules. DGANN distinguishes from previous models with those features: (1) It learns the local chemical environment encoding by graph attention mechanism on chemical bonds. Every initial edge message only flows into every message passing trajectory once. (2) The transformer blocks aggregate the global molecular representation from the local atomic encoding. (3) The position vectors and coordinates are used as inputs instead of distances. Our model has matched or outperformed most baseline graph neural networks on QM9 datasets even without thorough hyper-parameters searching. Moreover, this work suggests that models directly utilizing 3D coordinates can still reach high accuracies for molecule representation even without rotational and translational invariance incorporated.

## 1. Introduction

As is well-known, traditional experiments-driven new material development and drug discovery usually take a costly and lengthy process. Consequently, theoretical simulation has guiding significance to control the cost and accelerate the research and development process. Whereas, traditional specialized knowledge-inspired computational methods could also be quite time-consuming and only restricted to small scale screening with just a few candidates. For example, density functional theory[1,2] (DFT) and Hartree-Fock[3,4] (HF) methods are two widely used algorithms to solve the many-body Schrödinger equation with appropriate approximations. The DFT method usually takes a few minutes to hours, depending on the setups of the simulation such as the number of K-points in the irreducible Brillouin zone (IBZ), the cutoff energy for the plane-wave basis set, the size of the system, etc. Additionally, the more accurate but more complex ab initio algorithms like hybrid functionals[5] and GW approximation[6] are even more time-consuming. Therefore, high throughput molecular screening requires a more efficient method. Fortunately, the experiments and simulations in the literature generate a large amount of data which makes the deep learning method feasible.

Unlike image, video or text, the molecular structures are graph data that does not have the Euclidean properties and needs specialized method. The algorithms for the molecular structures include the traditional machine learning methods and the graph neural networks. Traditional methods are based on handcrafted features. The hand-engineered descriptors obtained by the domain expertise describe the chemical environment of each atom or the overall conformations. These descriptors are then usually processed by classical methods to obtain the atom-wise or structure-wise target. With the help of the well-designed descriptor as input, even a 2-layers neural network can fulfil high accuracy energy prediction for bulk silicon yet five orders of magnitude faster than DFT[7]. Support vector machine (SVM) method was used to identify the structural flow defects in glassy systems[8]. The Gaussian process regression (APR) developed potentials which we call Gaussian approximation potential (GAP), achieved high precision for carbon, silicon, germanium[9, 10]. The smooth overlap of atomic positions (SOAP) kernels also captured the intricate local chemical environments of molecules[11]. Clustering analysis identified the different grain boundaries of Cu[12]. Taking the Pauling electronegativity and ionic radii as input, neural networks with only two hidden layers can precisely predict the formation energy of the garnets and perovskites structures, which helps to classify the stable conformation[13]. The gradient-boosted regression trees (GBRT) was used to study the bandgap and heat of formation of perovskites with feature importance analyzed, facilitating the lead-free solar cell discovery[14].

In contrast to the classical machine learning methods, deep neural networks are not limited by the domain knowledge-driven feature engineering and work in an end-to-end fashion. Inspired by the remarkable success of deep neural networks in computer vision (CV) and natural language processing (NLP), graph neural networks were adopted for the tasks about the molecule and crystal structures. While many widely used graph neural networks in other domains only consider the node messages[15-24], the molecular networks make full use of the edge messages, mostly the pairwise distance. The convolutional neural network adapted from circular fingerprints predicted the toxicity and solubility with the corresponding molecular fragments identified[25]. Deep Tensor Neural Network (DTNN) uses the Hadamard product of source nodes message and the corresponding edge message (distance) in the interaction module to capture the interaction between the target nodes and all other nodes in the molecule[26]. The enn-s2s, on the other hand, adopted the discrete representation of the distance (distance bins) and featured with the edge network, multi-tower structure and the set-to-set readout function[27]. The SchNet as a variant of DTNN introduced the continuous-filter convolutions and the shifted softplus activation fuction[28]. Since then, many graph neural networks for quantum chemistries are devised with similar frameworks of SchNet and DTNN, e.g., HIP-NN[29], PhysNet[30] and MGCN[31]. The rotational covariant Cormorant is enlightened by the spherical tensors of quantum structures and captures the intramolecular interactions with comparative accuracies[32]. DimNet takes not only the distance of each edge but also the angle between edges into consideration. Ultrahigh precision was reached leveraging the spherical Bessel functions and spherical harmonic function[33]. Recent studies also applied the graph neural networks for transition states searching[34], chemical reactivity prediction[35, 36], water phase identification[37], the aqueous solubility of drug molecules[38], etc.

Following the guidance of the rotational, translational and permutation invariance, most models make full use of the atom pair distance (sometimes combined with the angle between edges). Those methods require a fully connected graph which takes atom pairs within the cutoff distance (usually includes all the atoms within the molecule) as edges. As a result, a fully connected graph which consists of N atoms and $N^2$ edges would have $N^2$ incoming messages for the distance-based methods. Furthermore, for the methods which also consider the angles between edges such as DimNet, the number of incoming messages would be $N^4$. The incoming messages would be computed separately in every layer and take much memory. Efforts have been made in searching for algorithms with high accuracy but without the complete graph. Hierarchical k-GNN architecture begins with the molecular graph whose edges are chemical bonds and allows the messages passing between k-dimension subgraph structures in deeper layers[39]. Even though the k-GNN is more potent than GNNs, the precision is still not comparative to the complete graph-based models, just like other GNNs[40].

To address this issue, we come up with the Directed Graph Attention Neural Network (DGANN) that does not require the complete graph and directly use the 3D coordinates instead of the distance as input in this paper. DGANN consists of interaction blocks, output blocks and transformer blocks. Like D-MPNN[41], we let the messages propagate through the directed chemical bonds but with attention mechanism for the interaction layers. Inspired by the position embedding in NLP[42], we take the position vectors of chemical bonds and coordinates of atoms as part of the inputs for the interaction and transformer block, respectively. The interaction and output blocks capture the local chemical environment within a few bonds of atoms. For previous models, the readout phase obtains the graph representation by direct summation or the set2set function of the nodes states[27]. Whereas, our framework adopts the more powerful transformer blocks[42] in the readout phase. The transformer blocks aggregate the global information of the molecules and obtain the high dimensional molecular fingerprints.

We further tested our model on the QM9[43] dataset for the 12 targets. Surprisingly, our framework not only achieves higher accuracy than DFT methods in all the targets but also outperforms the classical complete graph-based methods enn-s2s[27] and Cormorant[32] in most targets and reaches close precision in the remaining targets. The experiment shows that our model can directly comprehend the 3D geometrical information of molecules even without the rotational and translational invariance, which is different from our intuition.

## 2. Computational Method

Before introducing our model, we firstly recapitulate the classical scheme of the Message Passing Neural Networks[27]. The typical convolutional operation of most graph neural networks can be generalized into the message passing process and readout process. In other words, Message Passing Neural Networks is the base class to create different kinds of GNNs with various message propagation patterns.

For a GNN, the message passing phase runs T times and receives the messages from neighbouring atoms. At step t, a message-passing process includes the message function $M_t$ for edges $e_{vw}$ and the update function $U_t$ for nodes v[44]. The hidden states $h_v^{t+1}$ for the nodes v are obtained according to the following functions:

$$m_v^{t+1} = \sum_{w \in N(v)} M_t(h_v^t, h_w^t, e_{vw}) \qquad (1)$$

$$h_v^{t+1} = U_t(h_v^t, m_v^{t+1}) \qquad (2)$$

where $N_v$ denotes the collection of neighbour atoms of atom v, $M_t(h_v^t, h_w^t, e_{vw})$ is the messages calculated from the v and w nodes pair, $m_v^{t+1}$ is the messages obtained from the neighbour atoms of atoms v. The update functions obtain the new hidden states $h_v^{t+1}$ by combining the neighbourhood messages $m_v^{t+1}$ and the last-layer hidden states $h_v^t$.

The readout function R performs on all the nodes messages $h_v^T$ in graph G and predicts the output $\hat{y}$ for the target:

$$\hat{y} = R(\{h_v^T \mid v \in G\}) \qquad (3)$$

The aggregation schemes for the update phase have a variety of choices. Some models firstly obtain the vector representations of the graph (i.e., chemical fingerprint) and then use the Multilayer Perceptron (MLP) to predict. The molecular feature vector can be obtained by add[25], max-pooling[45] or set2set model of the last layer node states[27]. Other models apply species-wise MLPs on the last layer node states and sum over or take the average of the resulting scalar atomic contribution[26].

**2.1 The structure of DGANN**

The overall workflow of our framework is illustrated in Fig. 1. In this paper, we take aspirin as an example to elaborate on the mechanism of our work. Fig. 1a shows the simplified molecular-input line-entry system (SMILES) string and the conformation of aspirin. As presented by Fig. 1b, the Directed Graph Attention Neural Network (DGANN) includes three parts: the interaction blocks, an output block, and transformer blocks. The interaction block propagates the messages edge-wise along the directed path and captures the information of the path. The output block blends the atom messages and the connecting edges message, obtaining the representation of the local chemical environment. The transformer block aggregates the local messages and computes the global chemical fingerprint. All the blocks are based on the attention mechanism similar to the transformer.

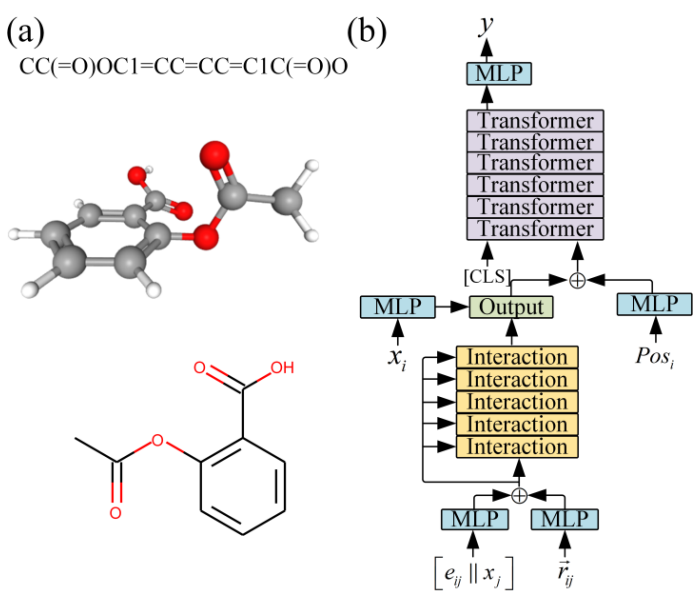

Figure 1. (a) The Smiles string and the conformation of aspirin. In this article, we use aspirin as an example to illustrate the working mechanism of our model. (b) The DGANN architecture. The ⊕ denotes element-wise add operation. The $e_{ij}$ represents the bond features for source node j to target node i. $x_i$ is the node feature for atom i. $\vec{r}_{ij} \in R^3$ is the position vector for the bond j → i. $Pos_i \in R^3$ is the Cartesian coordinates for atom i. [CLS] is a random vector used for the chemical fingerprint.

The interaction procedure for the edges is illustrated in Fig. 2. Fig. 2a shows the message passing direction along the edges around atom 2. For most models, messages interact through nodes. Whereas, like D-MPNN, we let the messages propagate along the edges. The input features for the edge j → i ∈ $R^{17}$ are the concatenation of the bond features $e_{ij}$ and the incoming atom features $x_j$. The bond features $e_{ij} \in R^4$ are the chemical graph edge representations[27] which are one-hot encodings of bond types (single, double, triple or aromatic). The atom feature $x_i \in R^{13}$ is same as which for MPNN[27]. The initial edge-wise hidden states $h_{ij}^0$ include messages from the edge features and the position vector as shown in Fig. 2b. Different from the position embedding in NLP domain, the position vector is continuous and should be processed by a neural network instead of direct embedding. The initial hidden states $h_{ij}^0$ are computed according to:

$$\begin{aligned} h_{ij}^0 = &(layernorm(W_{e2}gelu(layernorm(W_{e1}concat[e_{ij},x_j]))) \\ &+layernorm(W_{p3}gelu(layernorm(W_{p2}gelu(layernorm(W_{p1}\vec{r}_{ij})))))) / \sqrt{2} \end{aligned} \quad (4)$$

where the first and second term corresponds to the chemical edge feature and geometry information, respectively, $W_{e1} \in R^{d_{out} \times 17}, W_{e2} \in R^{d_{model} \times d_{out}}, W_{p1} \in R^{d_{out} \times 3}, W_{p2} \in R^{d_{out} \times d_{out}}, W_{p3} \in R^{d_{model} \times d_{out}}$, $d_{out}$ is the output dimension for a single attention head, $d_{model}$ is the model dimension which is equal to the product of $d_{out}$ and the number of attention heads. The $\sqrt{2}$ in Equation 4 is used to maintain the normal distribution of the initial hidden states magnitude.

The Gaussian error linear unit (GELU) is employed as the activation function. GELU has outmatched the RELU and ELU activations across the CV and NLP tasks in the previous studies[46].

$$gelu(x) = 0.5x(1+\tanh(\sqrt{\frac{2}{\pi}}(x+0.044715x^3))) \quad (5)$$

At each Interaction layer, queries matrix Q, keys matrix K and values matrix V are computed by the linear transformation without bias separately. The scaled dot-product of the query vector $\vec{q}_{ij}$ and the key vector $\vec{k}_{jk}$ is used as the attention score for the incoming edge jk to the receiving edge ij. A softmax function afterwards processes the attention scores among the relevant edges in Fig. 2d. The weighted sum of the value vectors $\vec{v}_{jk}$ is the output for the edge ij:

$$\vec{q}_{ij}, \vec{k}_{ij}^{l+1}, \vec{k}_{ij}^{0}, \vec{v}_{ij}^{l+1}, \vec{v}_{ij}^{0} = W_q h_{ij}^{0}, W_k h_{ij}^{l}, W_k h_{ij}^{0}, W_v h_{ij}^{l}, W_v h_{ij}^{0} \qquad (6)$$

$$\tilde{v}_{ij} = Attention(\vec{q}_{ij}, \vec{k}_{jk}, \vec{v}_{jk}) = \sum_{k \in N(j)} soft\max(\frac{\vec{q}_{ij} \cdot \vec{k}_{jk}}{\sqrt{d_{out}}}) \vec{v}_{jk} \qquad (7)$$

It should be noted that the input for the query matrix is the initial hidden state $\vec{h}_{ij}^{0}$, regardless of the depth of the layer l. Whereas, the input for the key matrix and the value matrix is the collection of the previous layer hidden states $\vec{h}_{ij}^{l}$ and the initial $\vec{h}_{ij}^{0}$. In other words, the hidden state at (l+1)-th layer $\vec{h}_{ij}^{l+1}$ is relevant to not only the incoming edges from the previous layer $\vec{h}_{jk}^{l}, k \in \{N(j)\backslash i\}$ but also the initial hidden states $\vec{h}_{ij}^{0}$. This structure ensures that every initial edge message only flows into every message passing trajectory once. Such a setting has two benefits: (1) Compared to the traditional GNNs based on interactions between nodes, it eliminates noises introduced by the repeated message travels between nodes. (2) If the depth of the layer is larger than the optimal value, the attention score would be pushed to zero for the redundant brand. Thus, it is capable of avoiding the vanishing gradient efficiently, like residual connection.

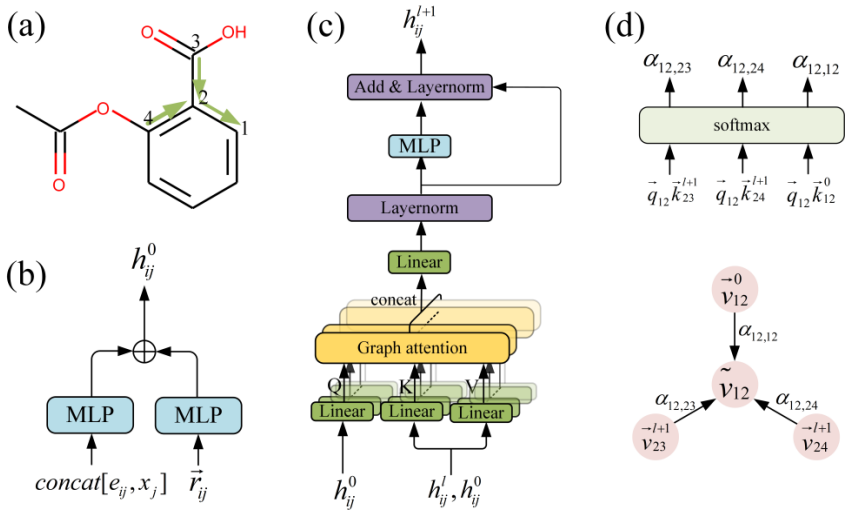

Figure 2. The mechanism of the interaction block. (a) The edge messages of edge 3 → 2 and 4 → 2 flow into 2 → 1, i.e. the hidden states $h^l_{24}$, $h^l_{23}$ and $h^0_{12}$ are used to update the $h^{l+1}_{12}$. (b) The initial hidden states $h^0_{ij}$ for edge j → i includes the messages from the bond feature $e_{ij} \in R^{13}$, the source atom feature $x_j \in R^4$ and the position vector of the bond $\vec{r}_{ij} \in R^3$. (c) The overall workflow of the interaction block. (d) The attention score is obtained by the scaled dot product of query with connecting keys followed by a softmax function. The attention scores then multiply with the corresponding values and aggregate to the target edge.

As denoted in Fig. 2c, the multi-head mechanism leverages multiple attention heads and takes the concatenation of the single head outputs. The following linear transformation and layer normalization mix the information from different subspaces to allow a more powerful representation:

$$\tilde{v}_{ij}^{multihead} = layernorm(W_0 concat(\tilde{v}_{ij}^1, \cdots \tilde{v}_{ij}^{nhead})) \tag{8}$$

where $W_0 \in R^{d_{model} \times d_{model}}$, $\tilde{v}_{ij}^n \in R^{d_{out}}$, nhead is the number of head.

The output of the multi-head attention $\tilde{v}_{ij}^{multihead}$ is then followed by the fully connected layer and residual connect to increase the nonlinearity of the model:

$$h_{ij}^{l+1} = layernorm(\tilde{v}^{multihead} + W_2 gelu(W_1 \tilde{v}^{multihead} + b_1)) \tag{9}$$

where $W_2 \in R^{d_{model} \times 2d_{model}}$, $W_1 \in R^{2d_{model} \times d_{model}}$, $b_1 \in R^{2d_{model}}$.

The output block blends the atomic information $x_i$ and the incoming edge messages $h_{ij}^l$ as shown in Fig. 3. The output $h_i^0$ of the block is the representation of the atomic local chemical environment and input for the following transformer block. The atom feature $x_i$ is firstly mapped into the higher-dimensional space $h_i^{init} \in R^{d_{model}}$ by an MLP with a similar structure to the first term of Equation 4. The output block and interaction block share the same structures with multi-head attention mechanisms in Fig. 3b. The input for the query vector is the high dimensional atom representation $h_i^{init}$. The corresponding key vector and the value vector are both obtained from the incoming last interaction layer edge hidden states $h_{ij}^l$ and the atom state $h_i^{init}$. The parameterized matrices $W_k$, $W_v$ are used for both $h_{ij}^l$ and $h_i^{init}$ because we wish to consider the influence of the incoming edges and node itself in a common mapping space.

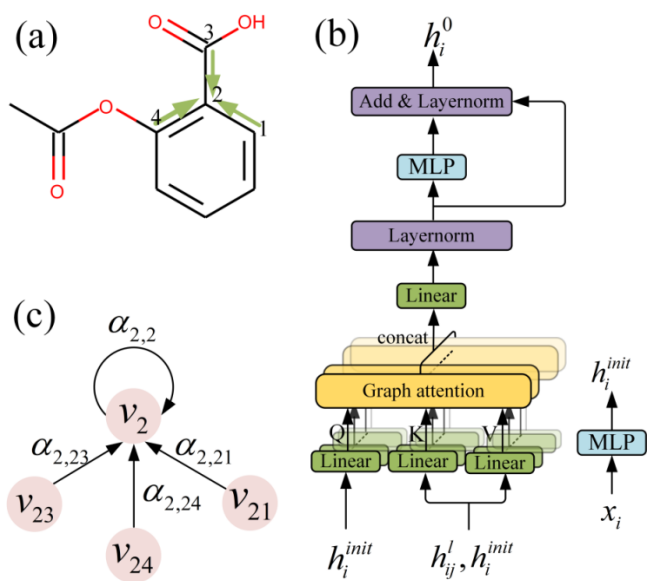

Figure 3. Illustration of the output block which aggregates the messages from the incoming edges and the node itself. The output is the node representation of the local chemical environment around atom 2. (a) The message passing direction around atom 2. (b) The workflow of the output block which is similar to the interaction block. (c) The attention mechanism illustration for atom 2.

The set2set model is capable of integrating the global feature relying on the attention mechanism [47] and performs excellently incorporated in enn-s2s[27]. Compared to the set2set, the transformer seems to be a more robust model utilizing the attention mechanism with great success in the field of NLP for the past few years[42]. Moreover, in a Kaggle competition to predict the atomic-wise scalar coupling, a Bert-based model that takes the position as part of the atom feature gets the third place in the competition. In this paper, we employ the transformer layer as the readout function to aggregate the global information from the atomic local chemical feature and obtain the vector representation for molecules.

The task to obtain the chemical fingerprint is similar to the sentiment analysis in NLP. We firstly pad the nodes hidden states of the output layer into a regular sequence for each molecule. A random normalized learnable vector is then concatenated to the beginning of every sequence as the initial representation of the global feature (as annotated by [CLS] in Fig. 4a). The corresponding first vector for the output of the last layer transformer is the chemical fingerprint.

We adopted a neural network for the position embedding of the continuous atomic Cartesian coordinate. Thus, the input of the first layer transformer can be expressed as:

$$h_i^{input} = (h_i^0 + layernorm(W_{pos3} gelu(layernorm(W_{pos2} gelu(layernorm(W_{pos1} Pos_i))))))/\sqrt{2} \quad \ldots\ldots(10)$$

where $W_{pos1} \in R^{d_{out} \times 3}, W_{pos2} \in R^{d_{out} \times d_{out}}, W_{pos3} \in R^{d_{model} \times d_{out}}$.

The chemical fingerprint is then followed by a 2-layer MLP to fit the target value.

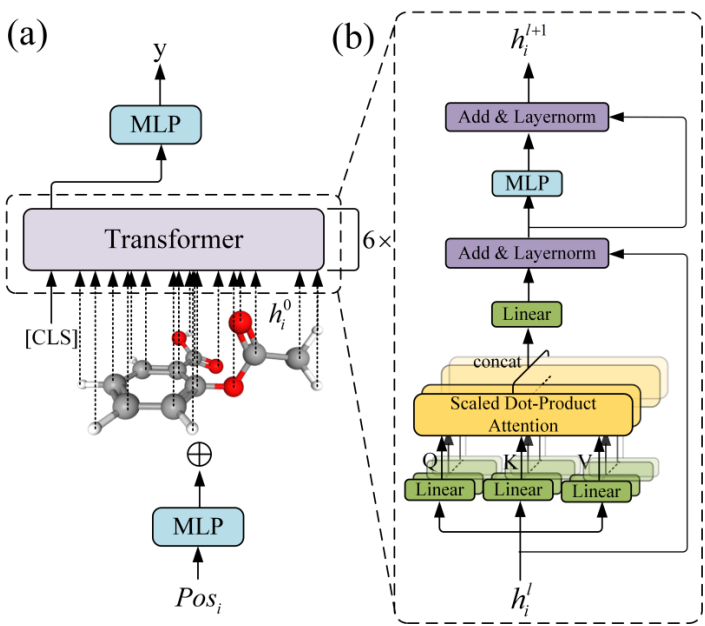

Figure 4. The illustration of the readout process. (a) The node-level local chemical representation $h_i^0$ and the position $Pos_i$ are processed by 6 transformer layers. The resulting global vector representation is then followed by a 2-layer MLP to learn the target value. (b) The structure of a transformer block.

## 2.2 Data augmentation

With the introduction of the position vector $\vec{r}_{ij}$ for the interaction block and the Cartesian coordinates $Pos_i$ for the transformer block, the rotation and translation invariance are not conserved in the model structure. As a result, the data augmentation would be of significant importance. In this work, we place the centre of the molecule at ordinate origin and apply the random rotation. We have also tried random translation for the molecules, yet this only makes the model difficult to converge.

## 2.3 Data Preprocessing

As we look into the range of value for the targets in QM9, the energy-related targets have much larger ranges than other targets. For example, the range of value for internal energy (eV) at 0K is $[-19444.385, -1101.488]$ in Fig. 5a, and the span is thousands of times more massive than most non-energy relates targets. Previous models handle this problem by either applying species-wise MLPs to every final atomic representation and take the summation as the molecule energy (e.g. SchNet[28]) or combining the method mentioned above with a hierarchical way (e.g. HIP[29]). Whereas, the normalized chemical fingerprint of fixed size is not enough to predict such subtle targets. Given that the atomic energy is highly sensitive to the atom species, we preprocess the target energy value by the least square method (LSM) for the targets of ZPVE, $U_0^{ATOM}$, $U^{ATOM}$, $H^{ATOM}$ and $G^{ATOM}$ according to:

$$y_{LSM} = X\theta \tag{11}$$

$$\theta = (X^T X)^{-1} X^T y \tag{12}$$

where $X \in R^{n_{train} \times 6}$, $\theta \in R^6$, $n_{train}$ is the size of the training set, the first 5 columns of X correspond to the number of atoms categorized to each species, the last column is padded by 1 corresponding to the bias.

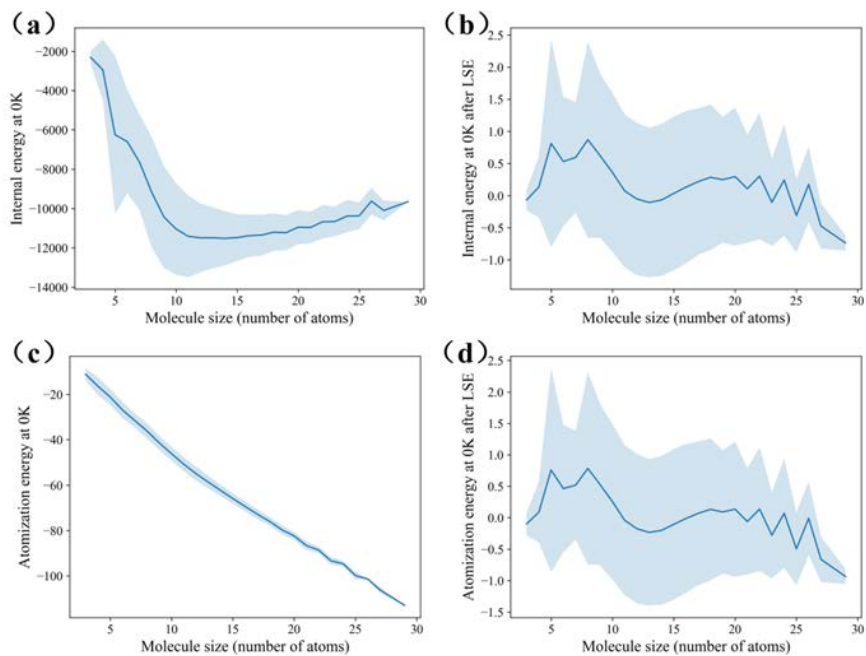

Figure 5. The confidence interval of molecular energy (eV) before and after LSM as a function of molecule size. (a) The internal energy at 0K. (b) The internal energy at 0K after the LSE. (c) The atomization energy at 0K. (d) The atomization energy at 0K after the LSE.

The role of the LSM is similar to the $0^{th}$ order output $\hat{E}_i^0$ of HIP. The nonlinearity of our model is then employed to predict the remaining energy. The LSM narrowed the target range of value significantly, as shown in Fig. 5. The range of values contracted 10000 and 100 times after the LSE for the internal energy and the atomization energy, respectively. Such operation also accelerates the convergence in the training process.

**2.4 Training**

For all the targets and models, we use the unified structures with the same hyperparameter settings. The dimension of our model and the number of heads are 512 and 8, respectively. There are 5 interaction blocks and 6 transformer blocks used. The Huber loss is employed in the training process. ADAM optimizer[48] is used to update the parameters for 600 epochs. The initial learning rate is 1e-5. Step learning rate decay is performed every 150 epochs with a decay rate of 0.5. We have also tried dropout and warm-up, but none of them seems to work. We trained 5 models for each target and selected the best model.

The QM9 dataset consists of 129433 molecules. We randomly split the dataset into the training set, validation set and test set as 0.9:0.05:0.05 with a batch size of 64. The validation set is used for model selection and early stopping, although the generalization error seldom increased before the final epoch in our training process. The mean average error (MAE) is used to evaluate our model.

## 3. Result and discussion

In this section, we test our DGANN on the QM9 dataset and compare it to the baseline models. The baseline models can be divided into two groups. One group operates on the graph where the edges are chemical bonds. Such models usually have relatively lower accuracy and complexity because of the negligence of long-range interaction. The other group employs the fully connected graph with the edges annotated by the atom-pair distance.

In Table 1, we show the mean average error (MAE) of different models on the QM9 dataset. For clarity, the best results are indicated in bold. Our DGANN outperforms other models in 7 out of 12 targets even without hyperparameters searching. For all the targets, DGANN archives lower MAE compared to the DFT method. Even though our model exceeds baseline models for the bandgap target, there is still room for improvement since the chemical accuracy has not been reached. The first two columns in Table 1 correspond to the non-complete graph methods. Although the DTNN performs precisely with a fully connected graph, the MAE is relatively large when we only take chemical bonds as input. Not surprisingly, the k-GNN performs better, considering the hierarchical model structure captures the intermediate-range interaction. Thanks to the fully connected graph, Cormorant and enn-s2s archive higher precision. The improvement can be explained by the long-range interaction directly included by the edges. The algorithm complexity is $O(N^2 d^2)$ for Cormorant, and $O(N^2 d^2/k)$ for enn-s2s (thanks to the tower structure of enn-s2s). It seems that for the previous models, the higher complexity is, the lower MAE would be. However, DGANN processes the short-range and long-range messages separately. The interaction block of DGANN only aggregates short-range messages from neighbour atoms with complexity equal to $O(|E|d^2)$. The transformer block process the global messages, and the corresponding time complexity is $O(N^2 d)$. Above all, the complexity of our model is lower than the Cormorant and enn-s2s counterparts.

In Table 2, we exemplify our model on several test set molecules for property prediction. The predicted property values are close to the target values in the bracket, which is satisfying. Du to that the bandgap is the most challenging but essential task for the property prediction, we analyzed the MAE for the bandgap prediction in Fig. 6. The molecule size distribution of the training set and test set are close with the random split used. Fig. 6c shows that the MAE for bandgap is insensitive to the molecule size. The chemical fingerprint can explain this insensitivity. In our model, the property value is computed following the fixed-length molecular vector representation that is normalized and irrelevant to the molecule size.

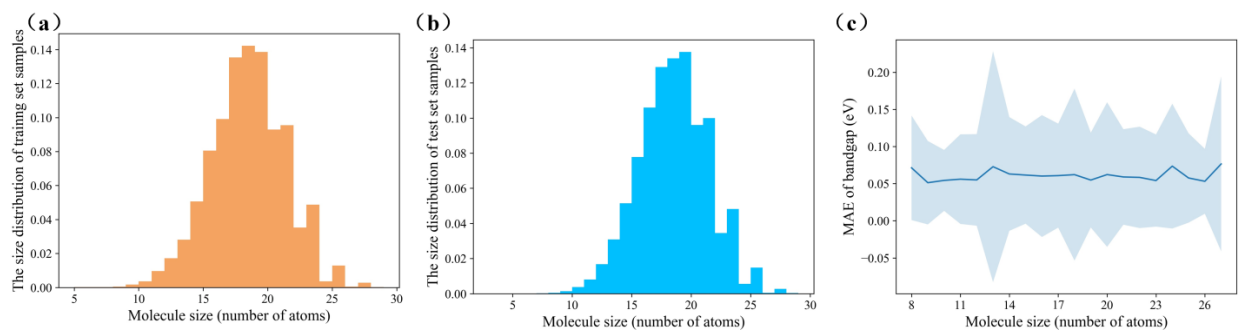

Figure 6. Panel (a) and (b) shows the molecule size distribution of the training set and test set, respectively, for bandgap prediction. (c) The MAE confidence interval of the bandgap for the test set as a function of the molecule size. The precision seems irrelevant to the number of atoms.

Fig. 7 shows the 2D chemical fingerprint for the bandgap target of all the molecules in the dataset. To facilitate the observation of the chemical fingerprint distribution in high dimensional space, we use the t-SNE method to reduce the dimensionality into 2D space. The colour in Fig. 7 indicates the target values. For clarity, we apply the cumulative distribution function to colour, since the bandgap values are not subject to the uniform distribution. Contrary to our intuition, molecules with close target values do not always cluster together as we can tell from the left panel of Fig.6. Furthermore, molecules with similar values sometimes can even be located at different corners of the chemical space. To further probe the chemical space, we magnify the vicinity of the molecule (the 6-mythlbicyclo[2.2.1]heptan-2-one structure is used for example in our paper) which we annotate as molecule 0 in the middle panel of Fig. 7. The numbers near the molecules in the dotted rectangle indicate the order of the distance to the molecule 0. For example, molecule N is the N-th nearest structure to molecule 0 in the high dimensional chemical space. The right panel of Fig. 7 lists the conformations of the molecules in the neighbourhood. The molecules around molecule 0 have similar structures. At this point, we can say that our model maps the 3D structures into a chemical space where molecules with similar structures and similar target values tend to be close.

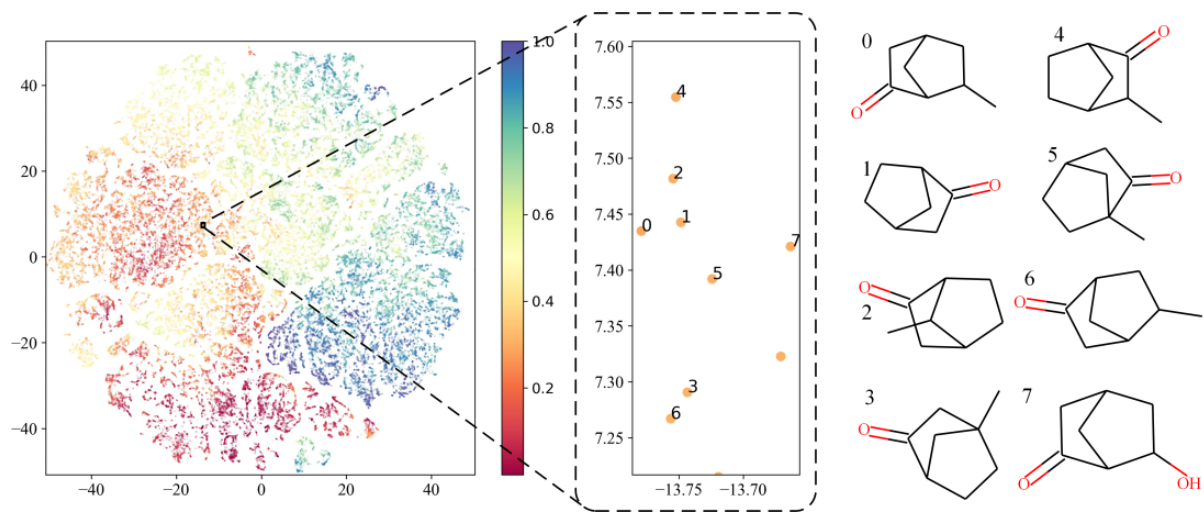

Figure 7. The 2D representation of the chemical fingerprint for the bandgap target. The low dimensional representations are obtained by t-SNE. The amplified region in the dashed box shows the nearest neighbouring molecules to the 6-methlbicyclo[2.2.1]heptan-2-one (also denoted as molecule 0 in the figure). The numbers near the molecules in the dotted rectangle indicate the ranking of the distance to the molecule 0. For example, molecule N is the N-th nearest structure to Molecule 0 in the high dimensional chemical space.

4. **Conclusions and future works**

In this work, we present the DGANN as a method for molecular property prediction. Different from the widely used complete graph-based models, our model does not operate on the distance annotated atom pair edges and only take the chemical bonds as edges. The model comprises the interaction, output and transformer blocks. The interaction and output blocks propagate messages along the edges and aggregate the local chemical environment around atoms. The transformer blocks aggregate the local atomic information and obtain the global vector representation for the molecules. Benefits from such settings, DGANN has lower time complexity compared to the complete graph-based methods. Moreover, the model directly takes the 3D information as input instead of the distance and does not conserve the rotational and translational invariance. Surprisingly, our model shows comparable or even better performance compared to the complete graph-based methods on QM9 dataset. In the chemical space, the DGANN makes the molecules with similar structures and target values cling close to each other. The MAE analysis of the bandgap shows that our model has no preference for the molecule size.

However, our model still has room for improvement, considering several drawbacks. Since the absence of rotation and translation invariance in the model structure, we have to increase model dimension and depth for better expression capacity. As a result, our model suffers from excessive parameters. The model is still computationally expensive, just like traditional complete graph-based methods, even though the time complexity is reduced theoretically. During the training process, we found our model sensitive to the randomly initialized parameters and believe it is relevant to the absence of rotation and translation invariance.

Table 1. Comparison of different models MAE on QM9 dataset. The best result is marked as bold for each target.

| Targets | DTNN (non-complete graph)[40] | k-GNN[39] | Comorant[32] | enn-s2s[27] | DGANN (ours) | DFT Error[49] | Chemical Accuracy[49] |
|---|---|---|---|---|---|---|---|
| $\mu(D)$ | 0.244 | 0.473 | 0.038 | **0.030** | 0.0340 | 0.1 | 0.1 |
| $\alpha(a_0^3)$ | 0.95 | 0.27 | 0.085 | 0.092 | **0.0820** | 0.4 | 0.1 |
| $\epsilon_{HOMO}(eV)$ | 0.1056 | 0.087 | 0.034 | 0.043 | **0.0375** | 2.0 | 0.043 |
| $\epsilon_{LOMO}(eV)$ | 0.1396 | 0.097 | 0.038 | **0.037** | 0.0380 | 2.6 | 0.043 |
| $\Delta\epsilon(eV)$ | 0.1796 | 0.133 | 0.061 | 0.068 | **0.0566** | 1.2 | 0.043 |
| $R^2(a_0^2)$ | 17.0 | 21.5 | 0.961 | **0.180** | 0.8562 | - | 1.2 |
| $ZPVE(eV)$ | 0.0468 | 0.0049 | 0.00203 | **0.0015** | 0.00189 | 0.0097 | 0.0012 |
| $U_0^{ATOM}(eV)$ | / | 0.9714 | 0.022 | 0.019 | **0.0156** | 0.1 | 0.043 |
| $U^{ATOM}(eV)$ | / | 2.9117 | 0.021 | 0.019 | **0.0159** | 0.1 | 0.043 |
| $H^{ATOM}(eV)$ | / | 1.1401 | 0.021 | 0.017 | **0.0164** | 0.1 | 0.043 |
| $G^{ATOM}(eV)$ | / | 1.2762 | 0.020 | 0.019 | **0.0163** | 0.1 | 0.043 |
| $C_v(cal\ mol^{-1}\ K^{-1})$ | 0.27 | 0.0944 | **0.026** | 0.040 | 0.0418 | 0.34 | 0.050 |

Table 2. Example of property prediction for structures in the test set. The values in the bracket indicate the target value.

| | OCCCC(=O)C#C 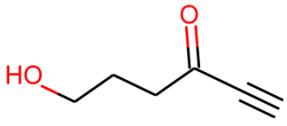 | CCc1cc(C)c(C)[nH]1 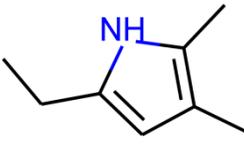 | C[C@@H](CCC#C)C1CC1 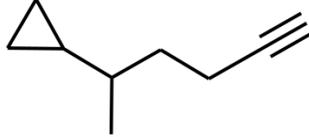 |
|---|---|---|---|
| $\mu(D)$ | 4.0387 (4.038) | 1.67219 (1.6797) | 0.87669 (0.872) |
| $\alpha(a_0^3)$ | 68.819 (68.81) | 92.82456 (92.84) | 92.44213 (92.45) |
| $\epsilon_{HOMO}(eV)$ | -7.2226 (-7.2165) | -4.94091 (-4.94431) | -6.95228 (-6.94979) |
| $\epsilon_{LOMO}(eV)$ | -1.5387 (-1.5347) | 1.37382 (1.37145) | 1.6069 (1.60819) |
| $\Delta\epsilon(eV)$ | 5.6665 (5.6817) | 6.3083 (6.31576) | 8.56018 (8.55798) |
| $R^2(a_0^2)$ | 1499.8997 (1500.3671) | 1498.52466 (1498.48682) | 1840.0127 (1839.92297) |
| $ZPVE(eV)$ | 3.4547 (3.4552) | 5.26299 (5.26298) | 5.53786 (5.53817) |
| $U_0^{ATOM}(eV)$ | 32.0997 (32.089) | 37.67077 (37.685) | 39.03687 (39.019) |
| $U^{ATOM}(eV)$ | -68.0598 (-68.0606) | -92.84554 (-92.84499) | -95.00114 (-94.99937) |
| $H^{ATOM}(eV)$ | -68.4196 (-68.421) | -93.41143 (-93.40646) | -95.59632 (-95.59881) |
| $G^{ATOM}(eV)$ | -68.8063 (-68.8066) | -93.95056 (-93.94627) | -96.15808 (-96.16431) |
| $C_v(cal\ mol^-$ | -63.5108 (-63.5089) | -86.49379 (-86.49363) | -88.36452 (-88.36382) |

**Notes and references**